\pdfoutput=1

\documentclass[11pt]{article}

\usepackage[preprint]{acl}

\usepackage{times}
\usepackage{latexsym}

\usepackage[T1]{fontenc}

\usepackage[utf8]{inputenc}
\usepackage{amsmath, amssymb, graphicx, tikz}
\usepackage{microtype}

\usepackage{inconsolata}
\usepackage{algorithm}
\usepackage{algpseudocode}
\usepackage{graphicx}
\usepackage{tikz}
\usepackage{authblk} 
\usetikzlibrary{quotes,backgrounds}
\title{Verbal Process Supervision Elicits Better Coding Agents}

\author[1,2]{\textbf{Hao-Yuan (Mark) Chen}}
\author[3]{\textbf{Cheng-Pong Huang}}
\author[3]{\textbf{Jui-Ming Yao}}

\affil[1]{Mindify AI, United States}
\affil[2]{University of London, United Kingdom}
\affil[3]{National Taiwan University of Science and Technology, Taiwan}

\affil[ ]{\texttt{\{mark, chengpong\}@mindifyai.dev}, \texttt{b11132009@mail.ntust.edu.tw}}

\begin{document}
\maketitle
\begin{abstract}
The emergence of large language models and their applications as AI agents have significantly advanced state-of-the-art code generation benchmarks, transforming modern software engineering tasks. However, even with test-time computed reasoning models, these systems still struggle with complex software engineering challenges. This work introduces CURA, a code understanding and reasoning agent system enhanced with verbal process supervision (VPS), achieving a 3.65\% improvement over baseline models on challenging benchmarks like BigCodeBench. Furthermore, CURA, when paired with the o3-mini model and VPS techniques, attains state-of-the-art performance. This work represents a step forward in integrating reasoning-driven architectures with LLM-based code generation, enabling agentic reasoning for language models to solve complex software engineering tasks.
\end{abstract}

\begin{quote}
\textit{"Journey before Destination."} \\
— Brandon Sanderson, \textit{The Way of Kings}
\end{quote}

\section{Introduction}
With the emergence of large-scale pre-trained language models \cite{brown2020languagemodelsfewshotlearners, touvron2023llama2openfoundation}, there has been a growing interest in their ability to act as autonomous agents capable of reasoning \cite{yao2023reactsynergizingreasoningacting}, planning \cite{wang2023planandsolvepromptingimprovingzeroshot}, reflection \cite{shinn2023reflexionlanguageagentsverbal} and executing complex tasks. In code generation, LLMs have demonstrated capable performance on various benchmarks. Yet, they often struggle with multi-step reasoning \cite{lightman2023letsverifystepstep}, debugging, and adapting to real execution feedback \cite{chen2022codetcodegenerationgenerated}. Traditional approaches rely on static datasets, limiting the model's capacity to refine its outputs dynamically\cite{shinn2023reflexionlanguageagentsverbal}. Therefore, this work introduces Code Generation and Reasoning Agent (CURA)—a reasoning framework enhanced by a novel verbal process supervision technique—to further improve the code generation capabilities of base reasoning and chat models. Verbal Process Supervision (VPS) is a novel supervision mechanism that enables language models to generate verbal process reward signals \cite{lightman2023letsverifystepstep}, guiding the reasoning process of the system, i.e., CURA.

The research explores whether iterative verbal process supervision, combined with agentic reasoning \cite{lightman2023letsverifystepstep} pipeline like CURA, can serve as a direct, fine-tuning-free method to reinforce model behavior and enhance reasoning capabilities through verbal reward signals to contrast the conventional reinforcement learning methods \cite{kumar2024traininglanguagemodelsselfcorrect}. To evaluate this, the study leverages BigCodeBench to assess large-scale frontier models \cite{deepseekai2025deepseekr1incentivizingreasoningcapability, openai2025competitiveprogramminglargereasoning}, examining whether this approach improves performance over a raw base model without agentic or iterative reasoning. Additionally, it introduces various chat models to determine whether verbal process supervision and the CURA architecture remain effective in smaller open-source models.

\section{Related Works}
In the related work section, the study investigates various notable works in the areas of code generation and agent reasoning framework to understand the frontier state of the areas while paving the foundation for the research to construct a concise yet precise definition of the research hypothesis.

\subsection{Agent Frameworks}
In recent works such as ReAct \cite{yao2023reactsynergizingreasoningacting} and Reflexion \cite{NEURIPS2023_1b44b878}, various attempts have been made to integrate agentic reasoning frameworks to enhance the performance of reasoning and coding models in solving complex software engineering tasks, including benchmarks like BigCodeBench \cite{zhuo2024bigcodebenchbenchmarkingcodegeneration} and HumanEval \cite{chen2021evaluatinglargelanguagemodels}. Reflexion, in particular, introduces a novel reinforcement learning approach by incorporating verbal reward signals into the learning loop, diverging from conventional policy gradient methods. This research serves as the theoretical foundation for the present study, which focuses on process-based supervision in reasoning frameworks such as ReAct and CURA. In contrast to Reflexion’s outcome-based approach, this study explores a complementary paradigm that emphasizes structured process-based reasoning supervision. 

\subsection{Code Generation Benchmarks}
Evaluating the code generation capabilities of large language models (LLMs) has been a key area of research, leading to the development of various benchmarks. Traditional benchmarks such as HumanEval \cite{chen2021evaluatinglargelanguagemodels} and MBPP \cite{austin2021programsynthesislargelanguage} primarily assess function-level code generation, where models generate self-contained functions given a natural language prompt.  

In contrast, BigCodeBench \cite{zhuo2024bigcodebenchbenchmarkingcodegeneration} evaluates LLMs’ ability to generate code that integrates multiple function calls across diverse libraries and domains. Unlike ClassEval \cite{du2023classevalmanuallycraftedbenchmarkevaluating}, which focuses on object-oriented programming, BigCodeBench assesses models on a broader range of practical programming tasks, including data analysis, networking, and system automation. It also introduces BigCodeBench-Instruct, a variant that reformulates task descriptions into natural language instructions to test LLMs' instruction-following capabilities.  

Empirical studies reveal that while LLMs perform well on isolated function calls, their accuracy deteriorates significantly when required to correctly compose multiple function calls. Both benchmarks highlight the limitations of current LLMs in handling complex, real-world programming tasks, underscoring the need for improved reasoning and compositional abilities in LLM-driven code generation.

\section{Methods}
The proposed work introduces CURA (Code Understanding and Reasoning Agent), a novel code generation framework incorporating verbal process supervision. This supervision mechanism enables language models to generate step-level reward signals, guiding them toward improved code-generation outcomes.

\subsection{CURA Architecture}
\begin{figure*}[t]
    \centering
    \includegraphics[width=\textwidth]{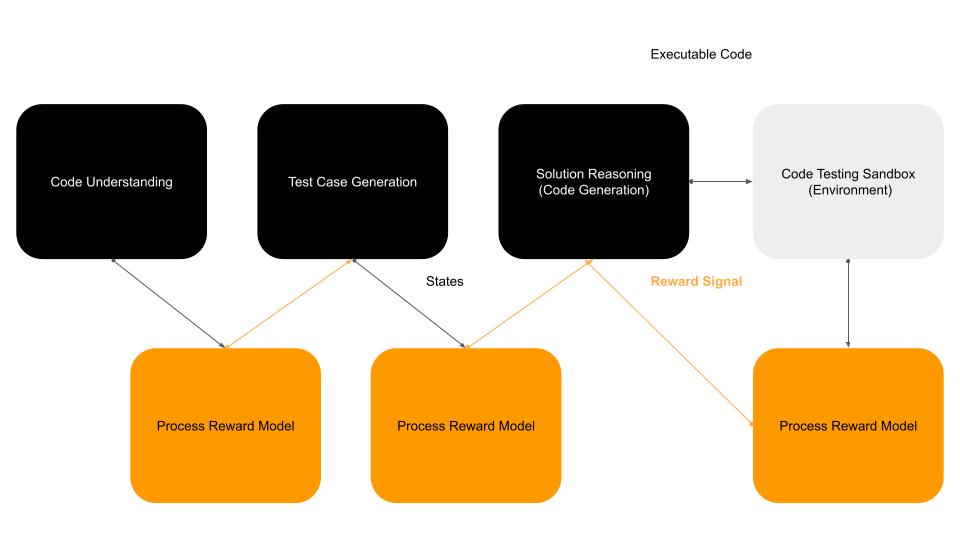}
    \caption{The CURA architecture: a process-supervised reasoning framework incorporating verbal reward signals.}
    \label{fig:cura-architecture}
\end{figure*}

The CURA architecture introduces an iterative, process-supervised reasoning framework for code generation, leveraging verbal reward signals to refine model behavior. The pipeline begins with code understanding, where the model interprets the given problem. Next, test case generation ensures the creation of diverse evaluation cases before transitioning to solution reasoning (code generation) to produce executable code. This code is then evaluated in a code-testing sandbox to verify correctness. At each stage, a process reward model provides state-based supervision, guiding the model through intermediate reasoning steps rather than relying solely on final execution results. Additionally, a reward signal from the code testing phase reinforces correct behavior, enabling fine-tuning-free improvements in reasoning and problem-solving. This structured, agentic approach enhances model performance by integrating iterative feedback and verbal process supervision throughout the entire reasoning pipeline (see Figure~\ref{fig:cura-architecture}) and algorithm \ref{alg:cura}.

\begin{algorithm}[t]
\caption{CURA Reasoning Framework}
\label{alg:cura}
\begin{algorithmic}[1]
\Require Problem statement, model parameters
\Ensure Generated and verified code solution
\State Initialize CURA reasoning framework
\While{recursion limit not reached}
    \State \textbf{Understanding:} Interpret the problem
    \State \textbf{Process Supervision:} Guide reasoning 
    \State \textbf{Test Generation:} Construct test cases
    \State \textbf{Process Supervision:} Guide reasoning 
    \State \textbf{Solution Reasoning:} Generate code
    \State \textbf{Process Supervision:} Guide reasoning 
    \State \textbf{Code Execution:} Run in sandbox
    \State \textbf{Verification:} Evaluate correctness
    \If{solution is correct}
        \State \Return final code solution
    \Else
        \State \textbf{Process Supervision:} Refine approach
    \EndIf
\EndWhile
\State \Return solution
\end{algorithmic}
\end{algorithm}

\subsection{Verbal Process Supervision (VPS)} 
In this study, verbal process supervision (VPS) enables iterative feedback at each step of the reasoning process. In contrast, the Reflexion agent \cite{shinn2023reflexionlanguageagentsverbal} refines its policy based on a verbal reward signal derived from the outcome. Overall, the VPS method comprises two models and an external environment. The first model is responsible for problem-solving and reasoning within the CURA pipeline, while the second model provides verbal rewards based on the current state of the processing pipeline. In code generation, the external environment acts as an isolated sandbox where executable code is rigorously tested. The resulting execution outcomes are then relayed to the reward model, which generates supervisory signals that steer and enhance the reasoning process of the code generation models, ultimately leading to improved final outputs.

\subsection{Prompt Engineering of Verbal Process Supervision (VPS)}
This structured verbal supervision approach ensures that each step is guided with natural language feedback, allowing iterative refinement before reaching the final solution. This methodology enhances reasoning capabilities, reduces hallucination errors, and enables process-level improvements in code generation tasks.

\begin{quote}
\textbf{Identity:} You are an expert AI assistant specializing in programmatic reasoning, problem decomposition, reflective reasoning, and solution verification.\\
\textbf{Context:} You are given a task description along with related outputs (such as task understanding, generated test cases, code, or error messages).\\
\textbf{Goal:} Provide a critique of the current output and suggest improvements if needed. You need to provide a detailed critique of the current output and suggest improvements to enhance the quality of the output.\\
\textbf{Task:} \{task\}\\
\textbf{Understanding:} \{task\_understanding\}\\
\textbf{Code:} \{code\}\\
\textbf{Test Code:} \{test\_code\}\\
\textbf{Error Message:} \{error\_message\}\\
\end{quote}

\section{Experiments}
This study validates the effectiveness of the CURA architecture combined with the VPS method on the BigCodeBench and HumanEval datasets. Using GPT-4o-mini and o3-mini models, the results demonstrate that CURA, when integrated with VPS, consistently improves performance across two different code generation tasks.

\subsection{BigCodeBench with Reasoning Model}
\begin{figure}[ht]
    \centering
    \includegraphics[width=1\linewidth]{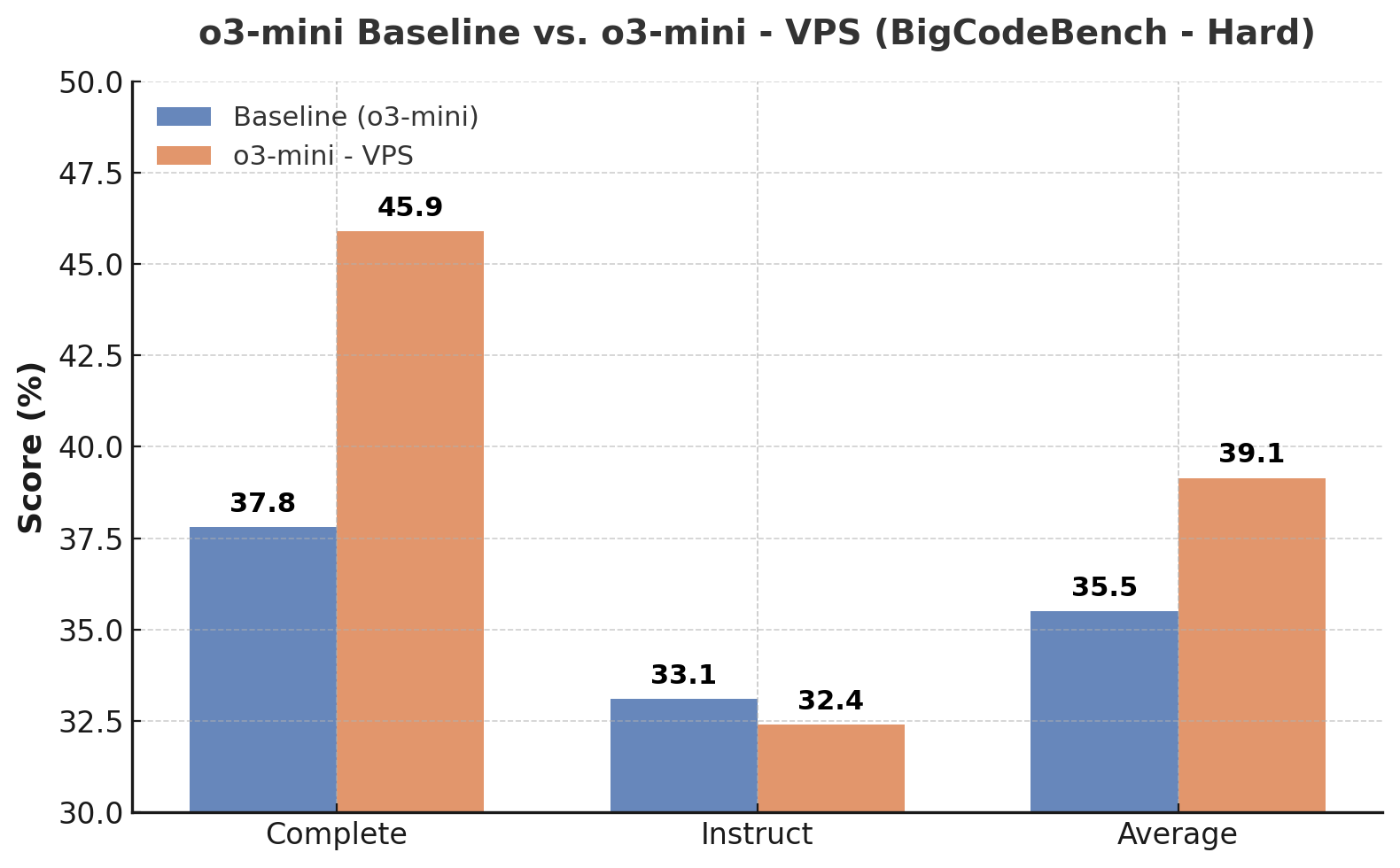}
    \caption{Comparison of o3-mini Baseline vs. o3-mini CURA with VPS on the BigCodeBench (Hard) dataset. 
    The y-axis shows the score (in \%), while the x-axis shows three different evaluation modes 
    (Complete, Instruct, and the Average of all modes). 
    Notice that o3-mini VPS shows an improvement in all categories, with the largest gain in the “Complete” mode.}
    \label{fig:o3-mini-vps-plot}
\end{figure}

The bar chart \ref{fig:o3-mini-vps-plot} compares the performance of the o3-mini Baseline and o3-mini - CURA with VPS on the BigCodeBench - Hard benchmark across three evaluation categories: Complete, Instruct, and Average. The vertical axis represents the score in percentage, ranging from 30\% to 50\%. The results indicate that the o3-mini-VPS model outperforms the o3-mini Baseline in the Complete and Average categories while slightly underperforming in the Instruct category. Specifically, in the Complete category, o3-mini - VPS achieves a score of 45.9\%, significantly higher than the o3-mini Baseline's 37.8\%. However, in the Instruct category, the baseline model attains a score of 33.1\%, which is marginally better than the VPS model's 32.4\%. When considering the Average score across all categories, o3-mini-VPS still demonstrates an overall improvement, scoring 39.1\% compared to the baseline's 35.5\%. These results suggest that the VPS-enhanced model provides substantial benefits in code completion tasks but may require further refinements for instruction-based tasks with pure natural language understanding.

\subsection{BigCodeBench with Various Chat Models on Different Temperatures}
\begin{figure}[ht]
    \centering
    \includegraphics[width=1\linewidth]{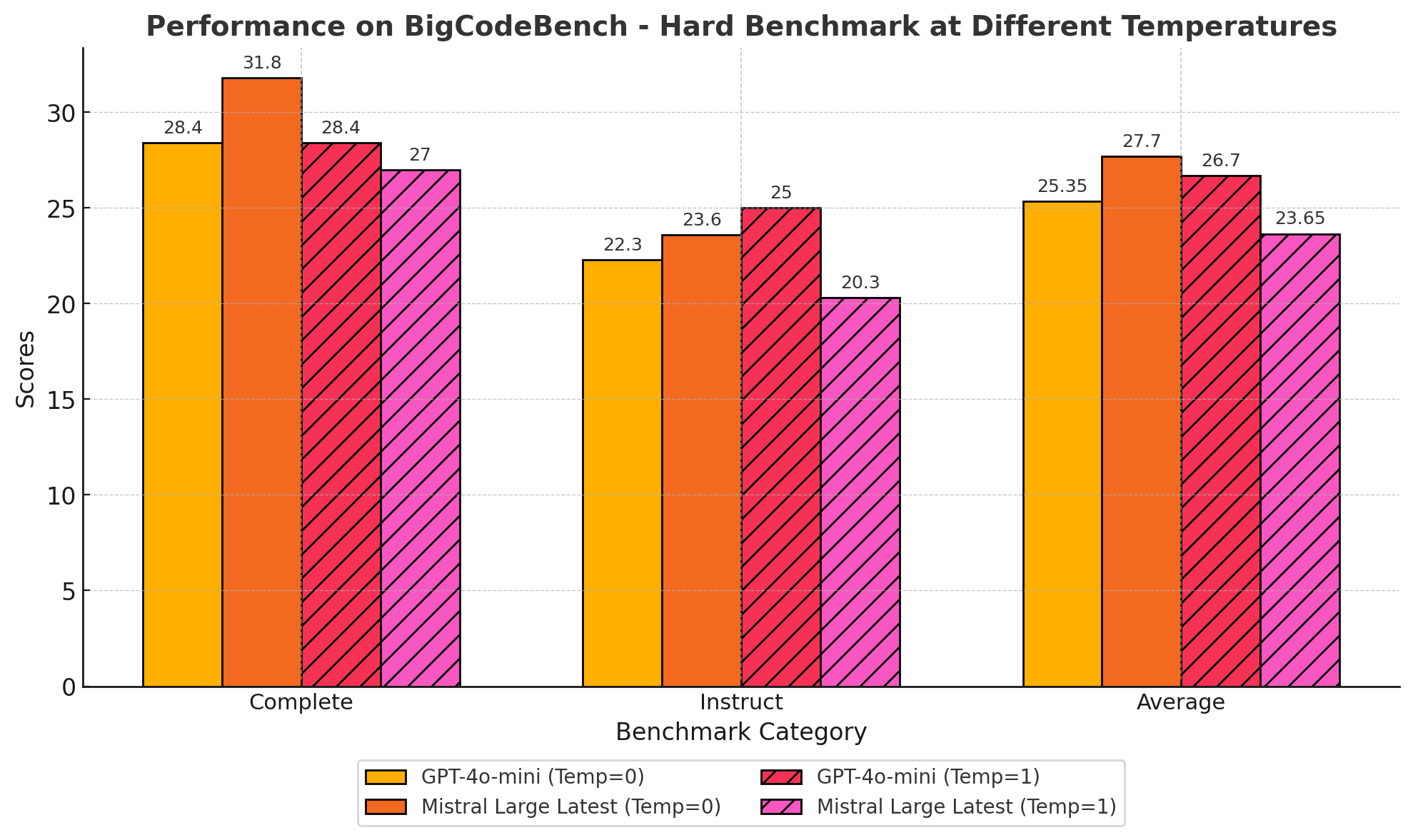}
    \caption{Performance comparison of GPT-4o-mini and Mistral Large Latest on the BigCodeBench using CURA architecture with VPS technique - Hard Benchmark across different temperature settings. The models are evaluated in three categories: Complete, Instruct, and Average. Results indicate that deterministic decoding (Temp=0) generally leads to higher scores, particularly in the Complete category where Mistral Large Latest outperforms other configurations. Increasing temperature (Temp=1) negatively impacts performance across all categories, highlighting the trade-offs between deterministic and stochastic decoding in code generation tasks.}
    \label{fig:performance-comparsion}
\end{figure}

The performance evaluation of different language models on the BigCodeBench - Hard Benchmark at varying temperature settings reveals key insights into their effectiveness in code generation tasks, which is shown in the figure \ref{fig:performance-comparsion}. The benchmark assesses models across three categories: Complete, which measures the ability to generate complete code segments; Instruct, which evaluates instruction-following capabilities; and Average, representing the overall performance. The comparison includes GPT-4o-mini and Mistral Large Latest, each evaluated at temperature 0 (deterministic setting) and temperature 1 (stochastic setting). The results indicate that models generally perform better at temperature 0, with Mistral Large Latest (Temp=0) achieving the highest score in the Complete category (31.8), surpassing GPT-4o-mini (Temp=0) (28.4). However, in the Instruct category, the gap is smaller, with GPT-4o-mini (Temp=0) scoring 22.3, while Mistral Large Latest (Temp=0) scores 23.6. The average performance follows a similar trend, where higher temperatures negatively impact scores, with Mistral Large Latest (Temp=1) demonstrating the lowest performance across all categories. This suggests that deterministic decoding strategies yield more reliable results, particularly in structured code generation tasks, whereas higher temperatures introduce randomness that degrades model effectiveness. These findings highlight the trade-offs between diversity and reliability in model outputs and suggest that Mistral Large Latest excels in deterministic code completion, while GPT-4o-mini remains competitive across categories.

\section{Discussion}
The results demonstrate that the CURA framework, when combined with verbal process supervision (VPS), significantly improves code generation performance across multiple benchmarks. The observed improvements are particularly evident in the BigCodeBench - Hard benchmark, where the CURA-enhanced models outperform their respective baselines in the Complete and Average evaluation categories. This indicates that process supervision effectively refines intermediate reasoning steps, leading to better overall performance. However, the slight underperformance in the Instruct category suggests that VPS may need further adaptation for tasks that heavily depend on direct instruction-following rather than iterative reasoning.

\subsection{Influence of Temperature Settings}
A key insight from the experiments is the influence of temperature settings on model performance. The results indicate that deterministic decoding (temperature = 0) generally yields more reliable outputs compared to stochastic decoding (temperature = 1). Mistral Large Latest achieves the highest scores in the Complete category when temperature is set to zero, while performance degrades as randomness increases. This suggests that structured code generation benefits from deterministic approaches, whereas higher temperatures introduce variability that can be detrimental to maintaining coherence and correctness.

The performance comparison between different models also reveals that while Mistral Large Latest excels in deterministic code completion, GPT-4o-mini maintains competitive performance across all categories. This indicates that while some models benefit more from structured reasoning, others balance instruction-following and code generation more effectively. The trade-offs between reliability and diversity in outputs suggest that future improvements could incorporate adaptive temperature strategies to optimize generation for different tasks.

\subsection{Challenges of CURA Method with VPS Technique}
The effectiveness of VPS in guiding intermediate steps also raises questions about its scalability to larger code-generation scenarios. The current setup applies VPS at a fine-grained level, providing immediate feedback at each stage of the reasoning process. While this enhances model refinement, it also introduces potential computational overhead. Future work should explore methods to balance the granularity of supervision with efficiency, potentially by leveraging hierarchical reward structures or selective intervention strategies.

\subsection{Opportunities of CURA Method with VPS Technique}
Another consideration is the broader applicability of CURA beyond code generation. The principles of verbal process supervision could be extended to other reasoning-intensive tasks, such as theorem proving, data analysis automation, or even multimodal reasoning. By incorporating process-level feedback rather than relying solely on final output evaluation, language models could develop more robust problem-solving capabilities across various domains.

\section{Conclusion}
This work introduces CURA, a novel reasoning architecture designed as a code understanding and reasoning agent, complemented by verbal process supervision (VPS) techniques. CURA leverages iterative, step-level VPS to guide large language models (LLMs), enabling them to significantly surpass their base performance. By integrating agentic reasoning with inference-time computation, this approach establishes a new paradigm for enhancing code generation and software engineering tasks. However, VPS remains constrained in its ability to precisely convey improvement directions to the code generation model. Future research will explore advanced prompting strategies and test-time computing techniques to further refine CURA and enhance the effectiveness of VPS.


\section*{Limitations}
While CURA with VPS demonstrates significant improvements in structured code generation, several limitations remain. One major challenge is the computational cost associated with iterative process supervision. Since VPS introduces multiple feedback loops, the overhead may slow down inference and limit the feasibility of real-time applications, especially in large-scale production environments.

Another limitation is the reliance on verbal feedback models, which may not always align with optimal reasoning strategies. Although verbal process supervision helps refine intermediate steps, it may occasionally introduce biases or incorrect guidance, particularly if the reward model is not well-calibrated. Further improvements in aligning VPS signals with human-expert feedback could enhance its effectiveness.

Additionally, the performance of CURA depends on the underlying language model's capabilities. While VPS helps refine reasoning, it does not fully mitigate the inherent limitations of pre-trained models, such as sensitivity to prompt variations and potential hallucinations in code generation. Addressing these challenges may require integrating stronger retrieval mechanisms or external knowledge bases to improve factual accuracy.

Finally, the adaptability of VPS across different domains remains an open question. Although the method is designed for code generation, its effectiveness in other complex reasoning tasks, such as mathematical theorem proving or multimodal reasoning, requires further investigation. Future work should explore domain-specific adaptations and evaluate VPS in a broader range of AI-assisted reasoning applications.


\section*{Acknowledgments}
The research was co-led by Mark Chen and Cheng-Pong Huang during their time at Mindify AI, whereas Jui-Ming Yao provided support and conducted an analysis of the empirical study of the work. 

\bibliography{custom} 

\begin{thebibliography}{15}
\providecommand{\natexlab}[1]{#1}

\bibitem[{Austin et~al.(2021)Austin, Odena, Nye, Bosma, Michalewski, Dohan, Jiang, Cai, Terry, Le, and Sutton}]{austin2021programsynthesislargelanguage}
Jacob Austin, Augustus Odena, Maxwell Nye, Maarten Bosma, Henryk Michalewski, David Dohan, Ellen Jiang, Carrie Cai, Michael Terry, Quoc Le, and Charles Sutton. 2021.
\newblock \href {https://arxiv.org/abs/2108.07732} {Program synthesis with large language models}.
\newblock \emph{Preprint}, arXiv:2108.07732.

\bibitem[{Brown et~al.(2020)Brown, Mann, Ryder, Subbiah, Kaplan, Dhariwal, Neelakantan, Shyam, Sastry, Askell, Agarwal, Herbert-Voss, Krueger, Henighan, Child, Ramesh, Ziegler, Wu, Winter, Hesse, Chen, Sigler, Litwin, Gray, Chess, Clark, Berner, McCandlish, Radford, Sutskever, and Amodei}]{brown2020languagemodelsfewshotlearners}
Tom~B. Brown, Benjamin Mann, Nick Ryder, Melanie Subbiah, Jared Kaplan, Prafulla Dhariwal, Arvind Neelakantan, Pranav Shyam, Girish Sastry, Amanda Askell, Sandhini Agarwal, Ariel Herbert-Voss, Gretchen Krueger, Tom Henighan, Rewon Child, Aditya Ramesh, Daniel~M. Ziegler, Jeffrey Wu, Clemens Winter, Christopher Hesse, Mark Chen, Eric Sigler, Mateusz Litwin, Scott Gray, Benjamin Chess, Jack Clark, Christopher Berner, Sam McCandlish, Alec Radford, Ilya Sutskever, and Dario Amodei. 2020.
\newblock \href {https://arxiv.org/abs/2005.14165} {Language models are few-shot learners}.
\newblock \emph{Preprint}, arXiv:2005.14165.

\bibitem[{Chen et~al.(2022)Chen, Zhang, Nguyen, Zan, Lin, Lou, and Chen}]{chen2022codetcodegenerationgenerated}
Bei Chen, Fengji Zhang, Anh Nguyen, Daoguang Zan, Zeqi Lin, Jian-Guang Lou, and Weizhu Chen. 2022.
\newblock \href {https://arxiv.org/abs/2207.10397} {Codet: Code generation with generated tests}.
\newblock \emph{Preprint}, arXiv:2207.10397.

\bibitem[{Chen et~al.(2021)Chen, Tworek, Jun, Yuan, de~Oliveira~Pinto, Kaplan, Edwards, Burda, Joseph, Brockman, Ray, Puri, Krueger, Petrov, Khlaaf, Sastry, Mishkin, Chan, Gray, Ryder, Pavlov, Power, Kaiser, Bavarian, Winter, Tillet, Such, Cummings, Plappert, Chantzis, Barnes, Herbert-Voss, Guss, Nichol, Paino, Tezak, Tang, Babuschkin, Balaji, Jain, Saunders, Hesse, Carr, Leike, Achiam, Misra, Morikawa, Radford, Knight, Brundage, Murati, Mayer, Welinder, McGrew, Amodei, McCandlish, Sutskever, and Zaremba}]{chen2021evaluatinglargelanguagemodels}
Mark Chen, Jerry Tworek, Heewoo Jun, Qiming Yuan, Henrique~Ponde de~Oliveira~Pinto, Jared Kaplan, Harri Edwards, Yuri Burda, Nicholas Joseph, Greg Brockman, Alex Ray, Raul Puri, Gretchen Krueger, Michael Petrov, Heidy Khlaaf, Girish Sastry, Pamela Mishkin, Brooke Chan, Scott Gray, Nick Ryder, Mikhail Pavlov, Alethea Power, Lukasz Kaiser, Mohammad Bavarian, Clemens Winter, Philippe Tillet, Felipe~Petroski Such, Dave Cummings, Matthias Plappert, Fotios Chantzis, Elizabeth Barnes, Ariel Herbert-Voss, William~Hebgen Guss, Alex Nichol, Alex Paino, Nikolas Tezak, Jie Tang, Igor Babuschkin, Suchir Balaji, Shantanu Jain, William Saunders, Christopher Hesse, Andrew~N. Carr, Jan Leike, Josh Achiam, Vedant Misra, Evan Morikawa, Alec Radford, Matthew Knight, Miles Brundage, Mira Murati, Katie Mayer, Peter Welinder, Bob McGrew, Dario Amodei, Sam McCandlish, Ilya Sutskever, and Wojciech Zaremba. 2021.
\newblock \href {https://arxiv.org/abs/2107.03374} {Evaluating large language models trained on code}.
\newblock \emph{Preprint}, arXiv:2107.03374.

\bibitem[{DeepSeek-AI et~al.(2025)DeepSeek-AI, Guo, Yang, Zhang, Song, Zhang, Xu, Zhu, Ma, Wang, Bi, Zhang, Yu, Wu, Wu, Gou, Shao, Li, Gao, Liu, Xue, Wang, Wu, Feng, Lu, Zhao, Deng, Zhang, Ruan, Dai, Chen, Ji, Li, Lin, Dai, Luo, Hao, Chen, Li, Zhang, Bao, Xu, Wang, Ding, Xin, Gao, Qu, Li, Guo, Li, Wang, Chen, Yuan, Qiu, Li, Cai, Ni, Liang, Chen, Dong, Hu, Gao, Guan, Huang, Yu, Wang, Zhang, Zhao, Wang, Zhang, Xu, Xia, Zhang, Zhang, Tang, Li, Wang, Li, Tian, Huang, Zhang, Wang, Chen, Du, Ge, Zhang, Pan, Wang, Chen, Jin, Chen, Lu, Zhou, Chen, Ye, Wang, Yu, Zhou, Pan, Li, Zhou, Wu, Ye, Yun, Pei, Sun, Wang, Zeng, Zhao, Liu, Liang, Gao, Yu, Zhang, Xiao, An, Liu, Wang, Chen, Nie, Cheng, Liu, Xie, Liu, Yang, Li, Su, Lin, Li, Jin, Shen, Chen, Sun, Wang, Song, Zhou, Wang, Shan, Li, Wang, Wei, Zhang, Xu, Li, Zhao, Sun, Wang, Yu, Zhang, Shi, Xiong, He, Piao, Wang, Tan, Ma, Liu, Guo, Ou, Wang, Gong, Zou, He, Xiong, Luo, You, Liu, Zhou, Zhu, Xu, Huang, Li, Zheng, Zhu, Ma, Tang, Zha, Yan, Ren, Ren, Sha, Fu, Xu, Xie, Zhang,
  Hao, Ma, Yan, Wu, Gu, Zhu, Liu, Li, Xie, Song, Pan, Huang, Xu, Zhang, and Zhang}]{deepseekai2025deepseekr1incentivizingreasoningcapability}
DeepSeek-AI, Daya Guo, Dejian Yang, Haowei Zhang, Junxiao Song, Ruoyu Zhang, Runxin Xu, Qihao Zhu, Shirong Ma, Peiyi Wang, Xiao Bi, Xiaokang Zhang, Xingkai Yu, Yu~Wu, Z.~F. Wu, Zhibin Gou, Zhihong Shao, Zhuoshu Li, Ziyi Gao, Aixin Liu, Bing Xue, Bingxuan Wang, Bochao Wu, Bei Feng, Chengda Lu, Chenggang Zhao, Chengqi Deng, Chenyu Zhang, Chong Ruan, Damai Dai, Deli Chen, Dongjie Ji, Erhang Li, Fangyun Lin, Fucong Dai, Fuli Luo, Guangbo Hao, Guanting Chen, Guowei Li, H.~Zhang, Han Bao, Hanwei Xu, Haocheng Wang, Honghui Ding, Huajian Xin, Huazuo Gao, Hui Qu, Hui Li, Jianzhong Guo, Jiashi Li, Jiawei Wang, Jingchang Chen, Jingyang Yuan, Junjie Qiu, Junlong Li, J.~L. Cai, Jiaqi Ni, Jian Liang, Jin Chen, Kai Dong, Kai Hu, Kaige Gao, Kang Guan, Kexin Huang, Kuai Yu, Lean Wang, Lecong Zhang, Liang Zhao, Litong Wang, Liyue Zhang, Lei Xu, Leyi Xia, Mingchuan Zhang, Minghua Zhang, Minghui Tang, Meng Li, Miaojun Wang, Mingming Li, Ning Tian, Panpan Huang, Peng Zhang, Qiancheng Wang, Qinyu Chen, Qiushi Du, Ruiqi Ge, Ruisong
  Zhang, Ruizhe Pan, Runji Wang, R.~J. Chen, R.~L. Jin, Ruyi Chen, Shanghao Lu, Shangyan Zhou, Shanhuang Chen, Shengfeng Ye, Shiyu Wang, Shuiping Yu, Shunfeng Zhou, Shuting Pan, S.~S. Li, Shuang Zhou, Shaoqing Wu, Shengfeng Ye, Tao Yun, Tian Pei, Tianyu Sun, T.~Wang, Wangding Zeng, Wanjia Zhao, Wen Liu, Wenfeng Liang, Wenjun Gao, Wenqin Yu, Wentao Zhang, W.~L. Xiao, Wei An, Xiaodong Liu, Xiaohan Wang, Xiaokang Chen, Xiaotao Nie, Xin Cheng, Xin Liu, Xin Xie, Xingchao Liu, Xinyu Yang, Xinyuan Li, Xuecheng Su, Xuheng Lin, X.~Q. Li, Xiangyue Jin, Xiaojin Shen, Xiaosha Chen, Xiaowen Sun, Xiaoxiang Wang, Xinnan Song, Xinyi Zhou, Xianzu Wang, Xinxia Shan, Y.~K. Li, Y.~Q. Wang, Y.~X. Wei, Yang Zhang, Yanhong Xu, Yao Li, Yao Zhao, Yaofeng Sun, Yaohui Wang, Yi~Yu, Yichao Zhang, Yifan Shi, Yiliang Xiong, Ying He, Yishi Piao, Yisong Wang, Yixuan Tan, Yiyang Ma, Yiyuan Liu, Yongqiang Guo, Yuan Ou, Yuduan Wang, Yue Gong, Yuheng Zou, Yujia He, Yunfan Xiong, Yuxiang Luo, Yuxiang You, Yuxuan Liu, Yuyang Zhou, Y.~X. Zhu,
  Yanhong Xu, Yanping Huang, Yaohui Li, Yi~Zheng, Yuchen Zhu, Yunxian Ma, Ying Tang, Yukun Zha, Yuting Yan, Z.~Z. Ren, Zehui Ren, Zhangli Sha, Zhe Fu, Zhean Xu, Zhenda Xie, Zhengyan Zhang, Zhewen Hao, Zhicheng Ma, Zhigang Yan, Zhiyu Wu, Zihui Gu, Zijia Zhu, Zijun Liu, Zilin Li, Ziwei Xie, Ziyang Song, Zizheng Pan, Zhen Huang, Zhipeng Xu, Zhongyu Zhang, and Zhen Zhang. 2025.
\newblock \href {https://arxiv.org/abs/2501.12948} {Deepseek-r1: Incentivizing reasoning capability in llms via reinforcement learning}.
\newblock \emph{Preprint}, arXiv:2501.12948.

\bibitem[{Du et~al.(2023)Du, Liu, Wang, Wang, Liu, Chen, Feng, Sha, Peng, and Lou}]{du2023classevalmanuallycraftedbenchmarkevaluating}
Xueying Du, Mingwei Liu, Kaixin Wang, Hanlin Wang, Junwei Liu, Yixuan Chen, Jiayi Feng, Chaofeng Sha, Xin Peng, and Yiling Lou. 2023.
\newblock \href {https://arxiv.org/abs/2308.01861} {Classeval: A manually-crafted benchmark for evaluating llms on class-level code generation}.
\newblock \emph{Preprint}, arXiv:2308.01861.

\bibitem[{Kumar et~al.(2024)Kumar, Zhuang, Agarwal, Su, Co-Reyes, Singh, Baumli, Iqbal, Bishop, Roelofs, Zhang, McKinney, Shrivastava, Paduraru, Tucker, Precup, Behbahani, and Faust}]{kumar2024traininglanguagemodelsselfcorrect}
Aviral Kumar, Vincent Zhuang, Rishabh Agarwal, Yi~Su, John~D Co-Reyes, Avi Singh, Kate Baumli, Shariq Iqbal, Colton Bishop, Rebecca Roelofs, Lei~M Zhang, Kay McKinney, Disha Shrivastava, Cosmin Paduraru, George Tucker, Doina Precup, Feryal Behbahani, and Aleksandra Faust. 2024.
\newblock \href {https://arxiv.org/abs/2409.12917} {Training language models to self-correct via reinforcement learning}.
\newblock \emph{Preprint}, arXiv:2409.12917.

\bibitem[{Lightman et~al.(2023)Lightman, Kosaraju, Burda, Edwards, Baker, Lee, Leike, Schulman, Sutskever, and Cobbe}]{lightman2023letsverifystepstep}
Hunter Lightman, Vineet Kosaraju, Yura Burda, Harri Edwards, Bowen Baker, Teddy Lee, Jan Leike, John Schulman, Ilya Sutskever, and Karl Cobbe. 2023.
\newblock \href {https://arxiv.org/abs/2305.20050} {Let's verify step by step}.
\newblock \emph{Preprint}, arXiv:2305.20050.

\bibitem[{OpenAI et~al.(2025)OpenAI, :, El-Kishky, Wei, Saraiva, Minaiev, Selsam, Dohan, Song, Lightman, Clavera, Pachocki, Tworek, Kuhn, Kaiser, Chen, Schwarzer, Rohaninejad, McAleese, o3~contributors, Mürk, Garg, Shu, Sidor, Kosaraju, and Zhou}]{openai2025competitiveprogramminglargereasoning}
OpenAI, :, Ahmed El-Kishky, Alexander Wei, Andre Saraiva, Borys Minaiev, Daniel Selsam, David Dohan, Francis Song, Hunter Lightman, Ignasi Clavera, Jakub Pachocki, Jerry Tworek, Lorenz Kuhn, Lukasz Kaiser, Mark Chen, Max Schwarzer, Mostafa Rohaninejad, Nat McAleese, o3~contributors, Oleg Mürk, Rhythm Garg, Rui Shu, Szymon Sidor, Vineet Kosaraju, and Wenda Zhou. 2025.
\newblock \href {https://arxiv.org/abs/2502.06807} {Competitive programming with large reasoning models}.
\newblock \emph{Preprint}, arXiv:2502.06807.

\bibitem[{Shinn et~al.(2023{\natexlab{a}})Shinn, Cassano, Berman, Gopinath, Narasimhan, and Yao}]{shinn2023reflexionlanguageagentsverbal}
Noah Shinn, Federico Cassano, Edward Berman, Ashwin Gopinath, Karthik Narasimhan, and Shunyu Yao. 2023{\natexlab{a}}.
\newblock \href {https://arxiv.org/abs/2303.11366} {Reflexion: Language agents with verbal reinforcement learning}.
\newblock \emph{Preprint}, arXiv:2303.11366.

\bibitem[{Shinn et~al.(2023{\natexlab{b}})Shinn, Cassano, Gopinath, Narasimhan, and Yao}]{NEURIPS2023_1b44b878}
Noah Shinn, Federico Cassano, Ashwin Gopinath, Karthik Narasimhan, and Shunyu Yao. 2023{\natexlab{b}}.
\newblock \href {https://proceedings.neurips.cc/paper_files/paper/2023/file/1b44b878bb782e6954cd888628510e90-Paper-Conference.pdf} {Reflexion: language agents with verbal reinforcement learning}.
\newblock In \emph{Advances in Neural Information Processing Systems}, volume~36, pages 8634--8652. Curran Associates, Inc.

\bibitem[{Touvron et~al.(2023)Touvron, Martin, Stone, Albert, Almahairi, Babaei, Bashlykov, Batra, Bhargava, Bhosale, Bikel, Blecher, Ferrer, Chen, Cucurull, Esiobu, Fernandes, Fu, Fu, Fuller, Gao, Goswami, Goyal, Hartshorn, Hosseini, Hou, Inan, Kardas, Kerkez, Khabsa, Kloumann, Korenev, Koura, Lachaux, Lavril, Lee, Liskovich, Lu, Mao, Martinet, Mihaylov, Mishra, Molybog, Nie, Poulton, Reizenstein, Rungta, Saladi, Schelten, Silva, Smith, Subramanian, Tan, Tang, Taylor, Williams, Kuan, Xu, Yan, Zarov, Zhang, Fan, Kambadur, Narang, Rodriguez, Stojnic, Edunov, and Scialom}]{touvron2023llama2openfoundation}
Hugo Touvron, Louis Martin, Kevin Stone, Peter Albert, Amjad Almahairi, Yasmine Babaei, Nikolay Bashlykov, Soumya Batra, Prajjwal Bhargava, Shruti Bhosale, Dan Bikel, Lukas Blecher, Cristian~Canton Ferrer, Moya Chen, Guillem Cucurull, David Esiobu, Jude Fernandes, Jeremy Fu, Wenyin Fu, Brian Fuller, Cynthia Gao, Vedanuj Goswami, Naman Goyal, Anthony Hartshorn, Saghar Hosseini, Rui Hou, Hakan Inan, Marcin Kardas, Viktor Kerkez, Madian Khabsa, Isabel Kloumann, Artem Korenev, Punit~Singh Koura, Marie-Anne Lachaux, Thibaut Lavril, Jenya Lee, Diana Liskovich, Yinghai Lu, Yuning Mao, Xavier Martinet, Todor Mihaylov, Pushkar Mishra, Igor Molybog, Yixin Nie, Andrew Poulton, Jeremy Reizenstein, Rashi Rungta, Kalyan Saladi, Alan Schelten, Ruan Silva, Eric~Michael Smith, Ranjan Subramanian, Xiaoqing~Ellen Tan, Binh Tang, Ross Taylor, Adina Williams, Jian~Xiang Kuan, Puxin Xu, Zheng Yan, Iliyan Zarov, Yuchen Zhang, Angela Fan, Melanie Kambadur, Sharan Narang, Aurelien Rodriguez, Robert Stojnic, Sergey Edunov, and Thomas
  Scialom. 2023.
\newblock \href {https://arxiv.org/abs/2307.09288} {Llama 2: Open foundation and fine-tuned chat models}.
\newblock \emph{Preprint}, arXiv:2307.09288.

\bibitem[{Wang et~al.(2023)Wang, Xu, Lan, Hu, Lan, Lee, and Lim}]{wang2023planandsolvepromptingimprovingzeroshot}
Lei Wang, Wanyu Xu, Yihuai Lan, Zhiqiang Hu, Yunshi Lan, Roy Ka-Wei Lee, and Ee-Peng Lim. 2023.
\newblock \href {https://arxiv.org/abs/2305.04091} {Plan-and-solve prompting: Improving zero-shot chain-of-thought reasoning by large language models}.
\newblock \emph{Preprint}, arXiv:2305.04091.

\bibitem[{Yao et~al.(2023)Yao, Zhao, Yu, Du, Shafran, Narasimhan, and Cao}]{yao2023reactsynergizingreasoningacting}
Shunyu Yao, Jeffrey Zhao, Dian Yu, Nan Du, Izhak Shafran, Karthik Narasimhan, and Yuan Cao. 2023.
\newblock \href {https://arxiv.org/abs/2210.03629} {React: Synergizing reasoning and acting in language models}.
\newblock \emph{Preprint}, arXiv:2210.03629.

\bibitem[{Zhuo et~al.(2024)Zhuo, Vu, Chim, Hu, Yu, Widyasari, Yusuf, Zhan, He, Paul, Brunner, Gong, Hoang, Zebaze, Hong, Li, Kaddour, Xu, Zhang, Yadav, Jain, Gu, Cheng, Liu, Liu, Wang, Lo, Hui, Muennighoff, Fried, Du, de~Vries, and Werra}]{zhuo2024bigcodebenchbenchmarkingcodegeneration}
Terry~Yue Zhuo, Minh~Chien Vu, Jenny Chim, Han Hu, Wenhao Yu, Ratnadira Widyasari, Imam Nur~Bani Yusuf, Haolan Zhan, Junda He, Indraneil Paul, Simon Brunner, Chen Gong, Thong Hoang, Armel~Randy Zebaze, Xiaoheng Hong, Wen-Ding Li, Jean Kaddour, Ming Xu, Zhihan Zhang, Prateek Yadav, Naman Jain, Alex Gu, Zhoujun Cheng, Jiawei Liu, Qian Liu, Zijian Wang, David Lo, Binyuan Hui, Niklas Muennighoff, Daniel Fried, Xiaoning Du, Harm de~Vries, and Leandro~Von Werra. 2024.
\newblock \href {https://arxiv.org/abs/2406.15877} {Bigcodebench: Benchmarking code generation with diverse function calls and complex instructions}.
\newblock \emph{Preprint}, arXiv:2406.15877.

\end{thebibliography}



\end{document}